# Early Prediction of Heart Disease Using PCA and Hybrid Genetic Algorithm with *k*-Means

Md. Touhidul Islam[1], Sanjida Reza Rafa[2], Md. Golam Kibria[3]

[1]Department of Computer Science and Engineering, City University, Dhaka, Bangladesh

[2]Department of Computer Science and Engineering, East West University, Dhaka, Bangladesh

[3]Department of Materials and Metallurgical Engineering, Bangladesh University of Engineering and Technology, Dhaka, Bangladesh

[1]touhid.cse@cityuniversity.edu.bd, [2]sanjidarezarafa@gmail.com, [3]kibria.buet004@gmail.com

*Abstract*—Worldwide research shows that millions of lives lost per year because of heart disease. The healthcare sector produces massive volumes of data on heart disease that are sadly not used to locate secret knowledge for successful decision making. One of the most important aspects at this moment is detecting heart disease at an early stage. Researchers have applied distinct techniques to the UCI Machine Learning heart disease dataset. Many researchers have tried to apply some complex techniques to this dataset, where detailed studies are still missing. In this paper, Principal Component Analysis (PCA) has been used to reduce attributes. Apart from a Hybrid genetic algorithm (HGA) with *k*-means used for final clustering. Typically, the *k*-means method is using for clustering the data. This type of clustering can get stuck in the local optima because this method is heuristic. We used the Hybrid Genetic Algorithm (HGA) for data clustering to avoid this problem. Our proposed method can predict early heart disease with an accuracy of 94.06%.

## I. Introduction

Health is considered a blessing that depends on proper care and several factors. Many civilizations developed for a few centuries and died away. Disease and decay overtook them.

Our civilizations are not free from health disease. This modern lifestyle has an impact on health disease which is increasing day by day. Among all other health diseases, heart disease is now a major concern as many people are losing their lives who are suffering from heart disease.

This increasing death rate can be lessened by early detection of heart disease. Many people are not conscious of the early detection of heart disease, though several healthcare industries are trying to detect heart disease at an early stage. However, it is still not noticed at an earlier stage. As a result, it becomes almost impossible to bear the cost of treatment. It is now a major concern among the researchers to detect heart disease at an early stage.

With the help of various data mining techniques, it is possible to detect heart disease at an early stage. The purpose of data mining is to extract information from a particular dataset and transform the information into a comprehensible structure for further use. The aim of our work is the early prediction of heart disease using PCA and hybrid genetic algorithm with *k*-means. PCA technique is used to reduce dimensionality. To make our working procedure easier, applying PCA our proposed algorithm converted the dataset attributes into two main attributes namely *PC1* and *PC2*.

A set of data points can be separated into groups or clusters using the Clustering technique [8]. The data points belonging to a cluster are identical and distinct from other clusters. *K*- means is a widely used data mining technique that is using for data clustering. But the nature of *k*-means is heuristic; for this reason, the method can be stuck in the local optima [8]. On the other hand, for better combinatorial optimization the best choice is Genetic Algorithms which is a metaheuristic type algorithm [9]. Our proposed method deterministically improved the final cluster by combining both techniques.

## II. Literature Review

Different research-based works have been done in the present years to find out the most preferable technique regarding heart disease prediction. In the paper [1] different data mining techniques have been analyzed by the authors to predict heart disease. The authors proposed an extended version of the model as a method which combines the generic algorithm for the selection of feature and fuzzy expert system for effective classification. After the experiment, the result of the Neural Network with 15 attributes has created an intensive impact compared to all other data mining techniques with 100% accuracy. The performance of the Decision Tree was also well with 15 attributes and the accuracy rate is 99.62%. Apart from this 99.2% efficiency has been shown by the Decision Tree in a combination of Genetic Algorithm and 6 attributes.

In the paper [2] authors have considered fourteen attributes such as age, sex, sugar, blood pressure which are extracted from a medical report to predict heart disease of the patient. The data mining techniques used in this research paper are *k*-means algorithms, MAFIA algorithm, Decision Tree classification for prediction of heart disease. It was concluded that Decision Tree has heart efficiency for the prediction of heart disease with fourteen attributes.

Another paper [3] has provided a survey of recent data mining techniques that are used in medical research specially to predict heart disease. Researchers have compared the performance of different data mining techniques and say that Decision Tree has performed well and Bayesian Classification also showing similar accuracy like Decision Tree. On the other hand, the performance of Neural Network, KNN, Classification based on clustering are not well compared to other methods.

In another paper [4] authors proposed an efficient genetic algorithm to predict heart disease with the backpropagation technique. KNN, Decision Tree, Naive Bayes are used as data mining classification techniques to analyze 13 attributes of a database of the healthcare industry. To predict the heart disease efficiently authors have presented an approach for extracting and fragmenting substantial forms from the heart attack data warehouses. The result shows that KNN has performed well compared to Decision Tree and Naive Bayes.

In the paper [5] authors have presented the development of a Neuro genetic model to predict coronary heart disease. The authors have used a multi-objective genetic algorithm to select a feature subset. The selected feature subset has been used for the prediction of the level of angiographic coronary heart disease with the help of a neural network. The result has shown 89.58% accuracy with the minimal feature set and also successfully reduced the complexity of the feature set.

Authors have presented a new hybrid model of Neural networks and genetic algorithms in the paper [6] to predict heart disease with the help of major risk factors. The major risk factors are -age, diabetes, hypertension, high cholesterol, tobacco smoking, alcohol intake, etc. The data of 50 patients with risk factors was collected to analyze and the result has shown 96.2% training accuracy along with 89% validation accuracy.

In the paper [7] to clustering data, authors have proposed a Hybrid Genetic Algorithm (HGA) with a different data set to avoid the problem that is generated by *k*-Means and *k*-Medoids. The authors also proposed a genetic encoding of the clustering problem and it is also mentioned that data points are separated into k clusters. Moreover, to calculate the fitness of the cluster Euclidean distance is used. Their proposed HGA methods showed a higher accuracy from 2.67% to 28.68% compared to other clustering accuracies that are mentioned in the related work.

III. PROPOSED METHOD

Principal Component Analysis, Hybrid Genetic Algorithm with *k*-Means two different kinds of data mining techniques are used for the early prediction of heart disease in the proposed method. The proposed method is shown in Fig. 2 and steps will be discussed below.

A. *Principal Component Analysis (PCA)*

Principal Component Analysis (PCA) is used to explain the variance-covariance structure of a set of variables through linear combinations [13]. It is used as a dimensionality-reduction technique. The first principal component has the highest variance possible. For each subsequent principal component, the variance decreases. UCI Machine Learning Repository heart disease dataset [11] has fourteen attributes; *target* is one of the attribute in the data set. For the convenience of work, we have turned thirteen attributes into two attributes i.e. *PC1* and *PC2* using the help of PCA.

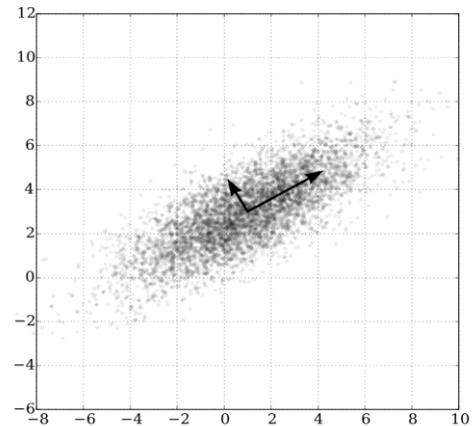

Fig. 1. Principal Component Analysis [12]

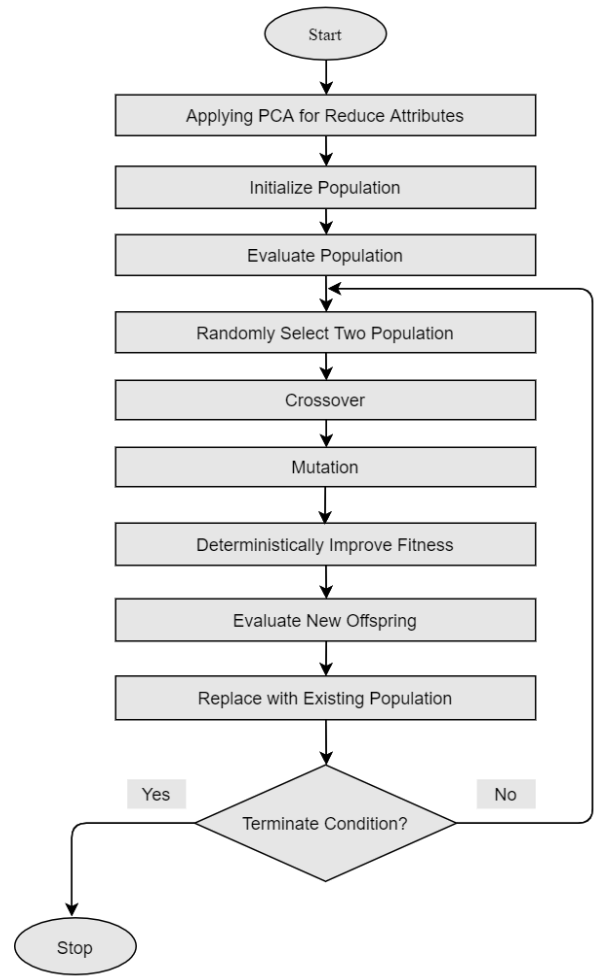

Fig. 2. Flow chart of the Proposed System

| Data Points | 1 | 2 | 3 | 4 | 5 |
|---|---|---|---|---|---|
| Target | 1 | 0 | 0 | 1 | 1 |

Target: Diagnosis of heart disease

Value 0: < 50% Diameter narrowing
(*Low risk of heart disease*)

Value 1: > 50% Diameter narrowing
(*High risk of heart disease*)

Fig. 3. Example Dataset

| Cluster \ Data Points | 1 | 2 | 3 | 4 | 5 |
|---|---|---|---|---|---|
| Low risk of heart disease | 0 | 1 | 0 | 0 | 0 |
| High risk of heart disease | 1 | 0 | 1 | 1 | 1 |

Fig. 4. Example data clustering. Data Point 2 belongs to cluster **Low risk of heart disease** and Data Points 1, 3, 4 and 5 belongs to cluster **High risk of heart disease**. Data Point 3 has been wrongly clustered.

*Chromosome 1*

| Cluster \ Data Points | 1 | 2 | 3 | 4 | 5 |
|---|---|---|---|---|---|
| Low risk of heart disease | 0 | 1 | 0 | 0 | 1 |
| High risk of heart disease | 1 | 0 | 1 | 1 | 0 |

*Chromosome 2*

| Cluster \ Data Points | 1 | 2 | 3 | 4 | 5 |
|---|---|---|---|---|---|
| Low risk of heart disease | 1 | 1 | 1 | 0 | 0 |
| High risk of heart disease | 0 | 0 | 0 | 1 | 1 |

*Offspring 1*

| Cluster \ Data Points | 1 | 2 | 3 | 4 | 5 |
|---|---|---|---|---|---|
| Low risk of heart disease | 0 | 1 | 0 | 0 | 0 |
| High risk of heart disease | 1 | 0 | 1 | 1 | 1 |

*Offspring 2*

| Cluster \ Data Points | 1 | 2 | 3 | 4 | 5 |
|---|---|---|---|---|---|
| Low risk of heart disease | 1 | 1 | 1 | 0 | 1 |
| High risk of heart disease | 0 | 0 | 0 | 1 | 0 |

Fig. 5. One Point Crossover

*Offspring 1*

| Cluster \ Data Points | 1 | 2 | 3 | 4 | 5 |
|---|---|---|---|---|---|
| Low risk of heart disease | 0 | 1 | 0 | 0 | 0 |
| High risk of heart disease | 1 | 0 | 1 | 1 | 1 |

*Offspring 1 after Mutation*

| Cluster \ Data Points | 1 | 2 | 3 | 4 | 5 |
|---|---|---|---|---|---|
| Low risk of heart disease | 0 | 0 | 0 | 0 | 1 |
| High risk of heart disease | 1 | 1 | 1 | 1 | 0 |

Fig. 6. Two Points Mutation for *Offspring 1*

|  | Offspring 2 | | | | |
|---|---|---|---|---|---|
| Cluster \ Data Points | 1 | 2 | 3 | 4 | 5 |
| Low risk of heart disease | 1 | 1 | 1 | 0 | 1 |
| High risk of heart disease | 0 | 0 | 0 | 1 | 0 |

|  | Offspring 2 after Mutation | | | | |
|---|---|---|---|---|---|
| Cluster \ Data Points | 1 | 2 | 3 | 4 | 5 |
| Low risk of heart disease | 0 | 1 | 1 | 1 | 1 |
| High risk of heart disease | 1 | 0 | 0 | 0 | 0 |

Fig. 7. Two points Mutation for *Offspring 2*

|  | Wrongly clustered | | | | |
|---|---|---|---|---|---|
| Cluster \ Data Points | 1 | 2 | 3 | 4 | 5 |
| Low risk of heart disease | 0 | 1 | 0 | 0 | 0 |
| High risk of heart disease | 1 | 0 | 1 | 1 | 1 |

|  | Improved cluster | | | | |
|---|---|---|---|---|---|
| Cluster \ Data Points | 1 | 2 | 3 | 4 | 5 |
| Low risk of heart disease | 0 | 1 | 1 | 0 | 0 |
| High risk of heart disease | 1 | 0 | 0 | 1 | 1 |

Fig. 8. Deterministically improvement of clustering quality. The Euclidean distance of data point 3 from the centroid of cluster *Low risk of heart disease* is less than the centroid of cluster *High risk of heart disease*. For data point 3, all possible reassignment is done in *Improved cluster*.

## B. Hybrid Genetic Algorithm with k-means

We introduce a genetic encoding technique for solving the clustering problem, where data points of every chromosome belong into two separate clusters, i.e. *Low risk of heart disease* and *High risk of heart disease*. An example dataset shows with five data points in Fig. 3 that data points are divided into two distinct categories. Data points 2, 3 belong to one category and 1, 4, 5 belong to another. Consider that every data points of the dataset have fourteen features; the *target* is one of the features in the data set. Without the *target* feature, the other thirteen features have been transformed into two features namely *PC1* and *PC2* by applying PCA. Suppose one of the parts of our proposed algorithm divided the chromosome into two clusters shows in Fig. 4. Data points 1 belongs cluster *Low risk of heart disease* and data points 1, 3, 4, 5 belongs cluster *High risk of heart disease*. Data point 3 of the dataset is wrongly clustered.

K-means method has been used to conclusive the fitness or nature of the clustering of a chromosome. On the other hand, Euclidean distance [8] has been used as *k*-means based HGA to measure the distance. The main objective of the HGA is to minimize the fitness so that the quality of clustering improves. Consider that the dataset has *n* data points. After randomly initializing chromosome value (binary 0 or 1); initially, for a chromosome, *p* data points from *n* data points belong to *Low risk of heart disease* and *q* data points belong to *High risk of heart disease*.

For the cluster of *Low risk of heart disease*,

$$Centroid, \quad p_{pc1} = \frac{\sum p_{PC1}}{p}, p_{pc2} = \frac{\sum p_{PC2}}{p}$$

$$fitness, \quad f_p = \sum \sqrt{(p_{PC1} - p_{pc1})^2 + (p_{PC2} - p_{pc2})^2} \quad (1)$$

Similarly, for the cluster of *High risk of heart disease*,

$$Centroid, \quad q_{pc1} = \frac{\sum q_{PC1}}{q}, q_{pc2} = \frac{\sum q_{PC2}}{q}$$

$$fitness, \quad f_q = \sum \sqrt{(q_{PC1} - q_{pc1})^2 + (q_{PC2} - q_{pc2})^2} \quad (2)$$

Final fitness of a chromosome,

$$f = f_p + f_q \quad (3)$$

Steady-state Genetic Algorithm [9] has been used and using deterministic improvement our proposed method made it hybrid to improve the clustering quality. At the beginning of the experiment, the original population of the chromosomes are randomly initialized. After experimenting with different population sizes, our algorithm has selected a final population size that has generated a better clustering. For calculating the fitness of the chromosomes several techniques have been used and those are described in (1), (2) and (3). Two parents are selected randomly to perform one-point crossover shows in Fig. 5. After crossover, two data points are randomly selected for mutation shows in Fig. 6 and Fig. 7.

For deterministic improvement consider that data point 3 in Fig. 4. From the figure, it indicates data point 3 is in that cluster which represents the *High risk of heart disease*. Using Euclidean distance, our algorithm determines the distance of data point 3 from the centroid of the two clusters. All possible reassignment of data points of two offspring has been performed by using this approach, shows in Fig. 8. The proposed algorithm determined the offspring's fitness using (1), (2) and (3) and find out the *maximum fitness* from the population. If *offspring1* fitness is less than *maximum fitness*, then the *maximum fitness* of the corresponding chromosome is replaced by *offspring1*. After all possible replacements, proposed algorithm again checked for *fitness2* where *maximum fitness* is already determined. Our proposed algorithm terminated the HGA when there is no improvement of minimum fitness values of chromosome after a consecutive doldrum period of *2 × Population size* generations. The solution to our work is the *minimum fitness* and corresponding chromosome.

## IV. DATASET AND RESEARCH FINDINGS

We implemented our proposed method using Python language with help of Jupyter Notebook and ran our proposed algorithm on an Asus laptop with Intel(R) Core (IM) i5-10210U CPU 2.11 GHz processor, 8GB RAM, and 64-bit Windows 10 Home OS. We have used UCI Machine Learning Repository [11] for the experiment heart disease dataset. Total of 303 data points and 14 attributes in the heart disease dataset. Attributes are age, sex, cp, trestbps, chol, fbs, restecg, thalach, exang, oldpeak, slope, ca, thal and target. The dataset has two data categories, Value 0: < 50% Diameter narrowing (*Low risk of heart disease*) and Value 1: > 50% Diameter narrowing (*High risk of heart disease*). *Low risk of heart disease* has 138 data points and *High risk of heart disease* has 165 data points.

Clustering quality is good or bad can be determined with a confusion matrix. It informs correctly and incorrectly clustered data points. For a *Population size 2500*, the clustering outcome of our proposed method has shown in TABLE I and Fig. 8 shows the final cluster applying our proposed method for the heart disease dataset. TABLE II evaluates the Clustering Accuracy, Clustering Error, Recall, Precision and *F*1 Score in percentage.

To evaluate the performance of our proposed method, we have used the following formulas for TABLE II:

$$Clustering\ Accuracy = \frac{TP + TN}{TP + FP + TN + FN} \times 100\%$$

$$Clustering\ Error = \frac{FP + FN}{P + N} \times 100\%$$

$$Recall = \frac{TP}{TP + FN} \times 100\%$$

$$Precision = \frac{TP}{TP + FP} \times 100\%$$

$$F1\ Score = \frac{2 \times Recall \times Precision}{Recall + Precision} \times 100\%$$

Here, separately TP, TN, FP, FN, P and N indicates True Positive, True Negative, False Positive, False Negative, Positive and Negative.

- **True Positive (TP):** Total number of correctly labeled positive predictive cases that are truly positive.
- **True Negative (TN):** Total number of correctly labeled negative predictive cases that are truly negative.
- **False Positive (FP):** Total number of truly negative cases that are labeled incorrectly as positive cases.
- **False Negative (FN):** Total number of truly positive cases that are labeled incorrectly as negative cases.
- **Positive(P):** Total number of real positive cases.
- **Negative(N):** Total number of real negative cases.

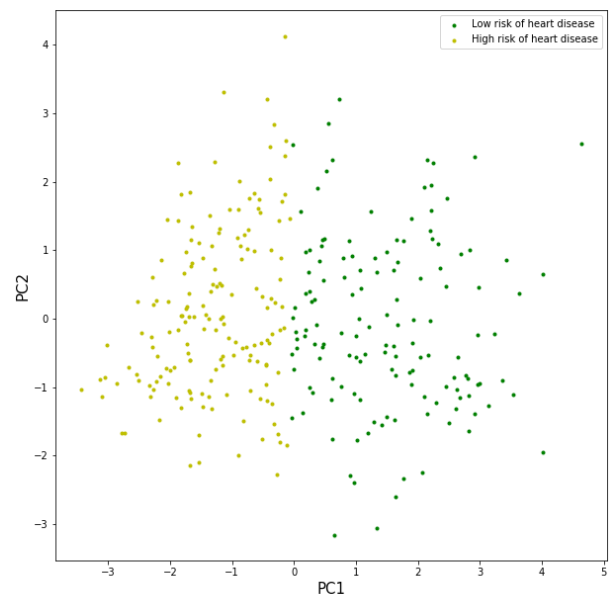

Fig. 9. The outcome of the proposed method

TABLE I. CONFUSION MATRIX

|  | | Predicted | | |
|---|---|---|---|---|
| | Cluster Type | Low risk of heart disease | High risk of heart disease | |
| Actual | Low risk of heart disease | 127 (TN) | 11 (FP) | N |
| | High risk of heart disease | 7 (FN) | 158 (TP) | P |

TABLE II. PERFORMANCE EVALUATION OF THE PROPOSED METHOD

| Clustering Accuracy | Clustering Error | Recall | Precision | F1 Score |
|---|---|---|---|---|
| 94.06% | 5.94% | 95.75% | 93.49% | 94.60% |

## V. CONCLUSION

Heart disease is considered one of the major threats to life and now it is a critical challenge to predict heart disease at an early stage in the area of clinical data analysis so that the death rate can be reduced. Our work intends to predict heart disease at an early stage. Our proposed method has worked in a few steps; each step does some work so that the final clustering quality is improved. Our method reduces the dimensionality of the dataset using PCA and combined the unsupervised heuristic k-means algorithm with metaheuristic Genetic Algorithms for better combinatorial optimization. After converging, our proposed algorithm deterministically improved the final clustering quality. The outcome reveals that these data mining techniques can predict heart disease early with an accuracy of 94.06%